\documentclass{article}

\PassOptionsToPackage{numbers, compress}{natbib}

\usepackage[numbers]{natbib}
\usepackage[preprint]{neurips_2026}

\usepackage[utf8]{inputenc} 
\usepackage[T1]{fontenc}    
\usepackage{hyperref}       
\usepackage{url}            
\usepackage{booktabs}       
\usepackage{amsfonts}       
\usepackage{nicefrac}       
\usepackage{microtype}      
\usepackage{xcolor}         
\usepackage{multirow}
\usepackage{graphicx}
\usepackage{xcolor}
\usepackage{animate}
\usepackage{wrapfig}

\usepackage{soul}        
\sethlcolor{yellow!30}   
\usepackage{booktabs} 
\usepackage{multirow} 
\usepackage{graphicx} 
\usepackage{tabularx}
\usepackage{makecell}
\usepackage{enumitem}
\usepackage{graphicx}
\usepackage{subcaption}

\usepackage{amsmath}
\usepackage{amssymb}
\usepackage{mathtools}
\usepackage{amsthm}
\usepackage{multirow}
\usepackage{graphicx}
\usepackage{animate}

\title{VISTA: Triplet-Supervised Video Style Transfer with Diffusion Transformers}

\author{%
  David S.~Hippocampus\thanks{Use footnote for providing further information
    about author (webpage, alternative address)---\emph{not} for acknowledging
    funding agencies.} \\
  Department of Computer Science\\
  Cranberry-Lemon University\\
  Pittsburgh, PA 15213 \\
  \texttt{hippo@cs.cranberry-lemon.edu} \\
}

\begin{document}

\author{
  Yiren Song$^{1}$\thanks{Equal contribution.} \quad
  Wangzi Yao$^{2}$\footnotemark[1] \quad
  Haofan Wang$^{3}$ \quad
  Mike Zheng Shou$^{1}$\thanks{Corresponding author.} \\
  \\
  $^{1}$Show Lab, National University of Singapore \\
  $^{2}$Institute of Automation, Chinese Academy of Sciences \\
  $^{3}$Lovart AI
}

\maketitle

\vspace{-9mm}

\begin{center}
    \centering
    \captionsetup{type=figure}
    \animategraphics[width=\linewidth]{5}{sec/image/teaser_q85/teaser-}{00001}{00010}
    \vspace{-4mm}
    \captionof{figure}{We introduce VISTA, a diffusion-transformer-based framework for video style transfer with principled triplet supervision. Leveraging the VISTA-1000 dataset with 1,000 styles and 12,000 motion-aligned triplets, VISTA disentangles style, content, and motion to achieve high-fidelity stylization with strong temporal consistency across complex motions. Readers can click and play the video clips in this figure using {\color{red}\textbf{Adobe Acrobat}}.}
\end{center}%

\begin{abstract}
Video style transfer aims to render videos in a target artistic style while preserving content, structure, and motion. While image stylization has advanced rapidly, video stylization remains challenging due to temporal inconsistency. Most existing methods stylize frames or keyframes and enforce consistency via heuristic temporal propagation, which is brittle under occlusions, disocclusions, and long-term motion, leading to drift and flickering artifacts. We argue that a fundamental bottleneck lies in the lack of large-scale triplet data and a principled training paradigm that jointly models and disentangles style, content, and motion.To address this, we introduce VISTA-1000, a synthetic dataset with 1,000 styles and motion-aligned triplets of style reference, clean video, and stylized video, and propose a diffusion-transformer-based in-context video style transfer framework with a lightweight style adapter for robust style extraction. Extensive experiments demonstrate SOTA performance in style fidelity, temporal consistency, and content preservation. Code is released at \href{https://github.com/showlab/VISTA}{https://github.com/showlab/VISTA}

\end{abstract}

\section{Introduction}

Video style transfer aims to transform an input video into a target artistic style while preserving its original content, structure, and motion. 
Despite rapid progress in image stylization, extending style transfer to videos remains challenging, particularly when long-term temporal coherence and complex motion are involved.
An effective video stylization system should not only maintain frame-wise visual quality, but also preserve motion dynamics and avoid temporal artifacts such as flickering, drifting textures, or structural distortion.

Most existing approaches address this challenge through a two-stage paradigm: applying image-based stylization to individual frames or keyframes, followed by heuristic temporal consistency enforcement using optical flow warping, feature propagation, or frame-wise regularization.
While such strategies can mitigate short-term flicker, they are inherently fragile under occlusions, disocclusions, and non-rigid motion.
Imperfect alignment errors accumulate over time, leading to unstable stylization in long videos.
These limitations suggest that temporal coherence cannot be reliably achieved through post-hoc propagation alone.

We argue that the core difficulty of video style transfer is not a lack of temporal smoothing techniques, but a more fundamental issue in how the problem is formulated and supervised.
Specifically, existing training setups fail to provide structured supervision that disentangles three essential factors: \emph{style appearance}, \emph{content structure}, and \emph{motion dynamics}.
Most datasets either contain only stylized videos without clean counterparts, or provide frame-level style signals without explicit motion alignment, forcing models to implicitly infer temporal consistency from weak or ambiguous supervision.
As a result, even powerful generative models tend to entangle appearance and motion, relying on brittle heuristics rather than learning motion-aligned stylization directly.

From a learning perspective, we argue that video style transfer should be formulated as a structured conditional generation problem with explicit triplet supervision.
Ideally, a model should be trained with triplets consisting of:
(1) a \emph{style reference image} that provides appearance cues,
(2) a \emph{clean input video} that specifies content and motion,
and (3) a \emph{stylized output video} that preserves the input motion and structure while adopting the target style.

Such triplets explicitly separate appearance from motion dynamics.
The clean and stylized videos share identical content and motion, while the style reference image provides stylistic cues with different semantics.
This structure enables the model to learn \emph{motion-aligned style transfer} directly from data, rather than enforcing temporal consistency through handcrafted constraints.
However, existing public benchmarks rarely provide such triplet supervision at scale.
Most datasets contain only a handful of styles, lack paired clean–stylized videos, or omit explicit style reference images, making it difficult to learn this disentanglement in a data-driven manner.

To address this limitation, we propose a data-driven learning paradigm that couples structured triplet supervision with a dedicated generative framework.
We introduce \textbf{VISTA-1000}, a large-scale synthetic dataset for motion-aligned video style transfer, containing 1,000 artistic styles and four semantically different but stylistically consistent videos per style.
Each style group provides paired clean--stylized videos with aligned motion and structure, where keyframes from one video naturally serve as style references for the others.
This construction yields 12,000 motion-aligned triplets and explicitly disentangles style appearance, content structure, and motion dynamics.


Building on this dataset, we propose \textbf{VISTA}, a diffusion-transformer-based video style transfer framework that learns motion-aligned stylization through in-context conditioning.
Instead of enforcing temporal coherence via optical flow or handcrafted regularization, our model jointly attends to style reference images, clean input videos, and denoising targets within a shared context.
This formulation allows temporal consistency and style fidelity to emerge naturally from paired supervision.
To further enhance style extraction from a single reference image, we introduce a lightweight Style Adapter with a two-stage training strategy that leverages large-scale image priors.
In addition, to improve the practicality of video-to-video stylization, we develop an efficient causal distillation pipeline that converts the bidirectional teacher into an autoregressive student with cross-chunk KV caching and further compresses inference into a few denoising steps.
This enables substantially faster video stylization while retaining the main visual behavior of the full model.

In summary, our contributions are threefold:
\begin{itemize}
    \item We revisit video style transfer from a learning perspective and argue that reliable temporal stylization requires \emph{structured triplet supervision} that explicitly disentangles style appearance, content structure, and motion dynamics.
    

    \item We publicly release \textbf{VISTA-1000}, a large-scale triplet dataset with 1,000 artistic styles, paired clean--stylized videos, and naturally formed style references for learning motion-aligned video style transfer.
    
    \item We propose a diffusion-transformer-based video-to-video style transfer framework with shared-context conditioning, a lightweight Style Adapter, and an efficient causal distillation strategy. The resulting system achieves strong style fidelity, temporal consistency, and content preservation, while supporting autoregressive and few-step accelerated inference without relying on optical flow or handcrafted temporal constraints.
\end{itemize}



\section{Related Works}
\label{sec:related}
\vspace{-2mm}

\subsection{Video Diffusion Models}
\vspace{-2mm}
With the rapid development of image and video generation models \cite{peebles2023scalable, wan2025wan, zheng2024open, jiang2025vace}, they are being used to create high-quality visual content and are widely applied in fields such as advertising, film special effects, game development, and animation production. Video generation models have evolved rapidly from early GAN-based approaches \cite{pan2017create}, UNet-based approaches \cite{guo2023animatediff, xu2024magicanimate} to today’s Diffusion Transformer architectures \cite{peebles2023scalable, wan2025wan, zheng2024open, jiang2025vace}. Modern Diffusion Transformers can generate high-quality, temporally coherent videos conditioned on text, images, or multi-modal inputs, enabling applications such as controllable video generation \cite{lin2025omnihuman, jiang2025vace, ma2025followyourclick, ma2025follow, ma2024followyouremoji, processpainter, song2025makeanything} and world modeling \cite{gao2025adaworld, song2025worldwander}. A significant portion
of the work, such as ControlNet \cite{zhang2023adding}, ControlVideo \cite{zhang2305controlvideo}, Composer \cite{huang2023composer}, VideoComposer \cite{wang2023videocomposer}, and SCEdit \cite{jiang2024scedit}, focuses
on single-condition editing and multi-condition composite editing based on temporal and spatial alignment conditions.

\subsection{Image Style Transfer}
\vspace{-2mm}
Image Style Transfer aims to transfer the visual appearance of a reference image onto a target content image while preserving the underlying semantic structure. Early neural style transfer methods rely on convolutional neural networks to align content and style feature statistics, including optimization-based approaches and feed-forward models based on feature normalization and reassembly (e.g., AdaIN\cite{huang2017arbitrary}, WCT\cite{li2017universal}).  With the advent of diffusion models, image style transfer has increasingly shifted toward conditional generation with large pretrained backbones. A line of work introduces lightweight adapters to inject reference image features into diffusion models through cross-attention or self-attention, enabling tuning-free or parameter-efficient stylization. Representative methods include IP-Adapter\cite{ye2023ip}, Style-Adapter\cite{wang2023styleadapter}, Instant Style\cite{wang2024instantstyle}, and Instant ID\cite{wang2024instantid}, which demonstrate strong flexibility in transferring diverse artistic styles. Another line of work directly optimizes the parameters of vector graphics or differentiable primitives with CLIP-based objectives to produce stylized visual effects \cite{song2022cliptexture, song2023clipvg, song2022clipfont, schaldenbrand2022styleclipdraw}. More recently, in the Diffusion Transformer era, image stylization has moved toward unified and consistency-aware modeling, where style representations are learned in a more generalizable manner across images and tasks. Methods such as OmniConsistency\cite{song2025omniconsistency} exemplify this trend, highlighting the potential of diffusion transformers for scalable and robust image style transfer.

\subsection{Video Stylization and Editing}
\vspace{-2mm}
Early video style transfer methods extend image stylization to videos by applying frame-wise processing followed by temporal consistency enforcement via optical flow warping, feature propagation, or handcrafted regularization~\cite{huang2017real,gao2020fast}.
While these techniques can mitigate short-term flicker, they are brittle under occlusions, non-rigid motion, and long temporal horizons, where alignment errors accumulate and cause texture drifting or structural instability, suggesting that post-hoc propagation alone is insufficient for reliable temporal coherence~\cite{ruder2016artistic,wang2020consistent}. 
Recent advances in diffusion-based foundation models have also substantially promoted image and video editing, enabling users to manipulate visual content through text instructions, reference images, masks, structural controls, or in-context examples~\cite{batifol2025flux, yan2025eedit, guo2025any2anytryon, gong2025relationadapter, liu2025omnirefiner, ouyang2025consistency, chen2026edittransfer++}.
Building on these generative priors, recent diffusion-based methods have advanced video stylization and editing by casting them as video-to-video generation, improving visual fidelity and temporal stability.
Representative methods introduce structured control signals or decoupled appearance--structure representations (e.g., CCEdit~\cite{feng2024ccedit}), or unify video creation and editing within a single diffusion framework (e.g., VACE~\cite{jiang2025vace}).
Other works explore training-free or inversion-based pipelines for localized or long video stylization, leveraging mask propagation, sliding-window processing, or attention-based control~\cite{qi2023fatezero,geyer2023tokenflow}.
However, most existing diffusion-based methods still rely on heuristic alignment or implicit temporal constraints, and few leverage large-scale structured supervision that explicitly disentangles style appearance, content structure, and motion dynamics~\cite{qin2024instructvid2vid}.
In contrast, our work learns motion-aligned video style transfer directly from triplet supervision, allowing temporal consistency and stylistic coherence to emerge from data rather than heuristic propagation.

\section{Method}
\label{sec:method}
\vspace{-2mm}

\subsection{Problem Formulation}
\vspace{-2mm}
\label{sec:problem_formulation}

We study video style transfer under triplet supervision.
Each training sample consists of a style reference image $I_s$, a clean input (source)
video $V_c = \{x_t\}_{t=1}^{T}$, and a stylized target video
$V_s = \{y_t\}_{t=1}^{T}$.
The source and target videos share identical content and motion, while the reference
image provides appearance cues.

Our goal is to transfer the style of $I_s$ onto $V_c$ while preserving the content
structure and motion dynamics of the source video.
Formally, we aim to model the conditional distribution
\vspace{-0.2em}
\begin{equation}
V_s \sim p(\,\cdot \mid I_s, V_c\,).
\end{equation}
\vspace{-0.2em}
To this end, we formulate video stylization as a token-level conditional generation
problem.
The reference image, the source video, and the target video are encoded as style image
tokens $T_s$, source video tokens $T_c$, and target video tokens $T_d$, respectively.
These tokens form a shared context that serves as the input to our generative model.

\subsection{Shared-Context Diffusion for Motion-Aligned Stylization}
\label{sec:shared_context}

We instantiate the conditional distribution $p(V_s \mid I_s, V_c)$ using a Diffusion
Transformer (DiT) operating on the shared token context defined in
Sec.~\ref{sec:problem_formulation}.
All tokens are concatenated and processed by a transformer with bidirectional
self-attention.

\begin{figure*}[t]
\centering
\includegraphics[width=\linewidth]{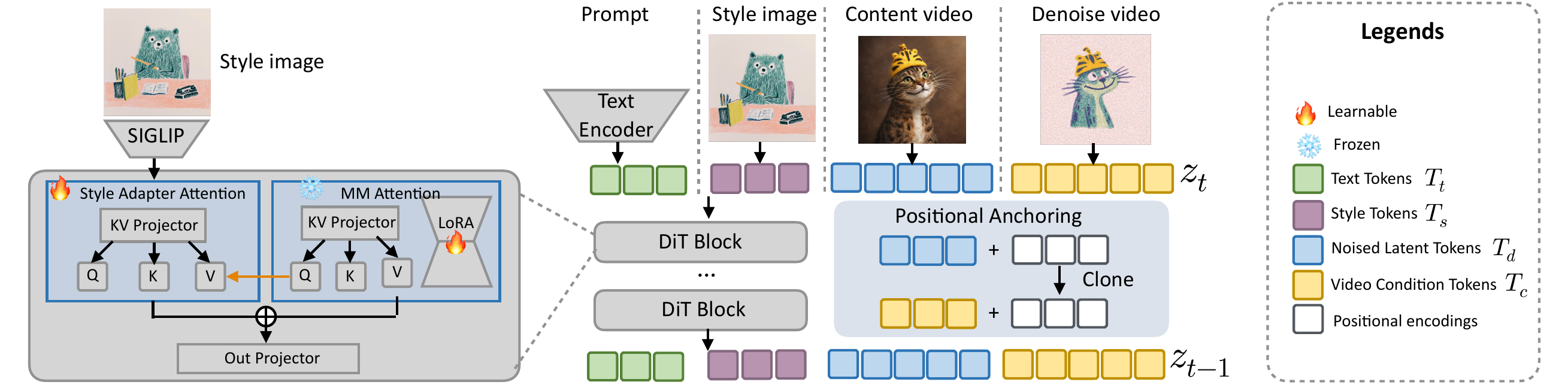}
\captionof{figure}{Overview of the VISTA framework. VISTA utilizes a shared-context Diffusion Transformer to jointly model style, content, and motion. It integrates a Style Adapter (left) for appearance control and Positional Anchoring (right) for precise spatiotemporal alignment, ensuring temporal consistency without optical flow or handcrafted heuristics.}
\vspace*{-1em}
\label{fig:structure}
\end{figure*}

Unlike conventional conditional diffusion models that inject conditions via
unidirectional cross-attention, our shared-context design places style, content, and
motion tokens within a single attention space, as shown in Figure \ref{fig:structure}.
This allows denoising tokens $T_d$ to jointly attend to appearance cues from $T_s$ and
spatiotemporal information from $T_c$ across all transformer layers.
As a result, the model can associate style transformation directly with motion patterns
observed in the source video, rather than propagating style independently across frames.

\paragraph{Diffusion Objective.}
Let $T_{d,0}$ denote the latent tokens corresponding to the ground-truth stylized
video $V_s$.
At diffusion timestep $t$, Gaussian noise is added as
\begin{equation}
T_{d,t} = \sqrt{\bar{\alpha}_t}\, T_{d,0} + \sqrt{1 - \bar{\alpha}_t}\, \epsilon,
\quad \epsilon \sim \mathcal{N}(0, I),
\end{equation}
where $\bar{\alpha}_t$ follows a predefined noise schedule.
The model is trained to predict the injected noise conditioned on the shared context:
\begin{equation}
\mathcal{L}_{\text{diff}} =
\mathbb{E}_{t,\epsilon}
\left[
\left\|
\epsilon -
\epsilon_{\theta}\!\left(T_s,\, T_c,\, T_{d,t},\, t\right)
\right\|_2^2
\right].
\end{equation}

By conditioning the denoising process on the entire source video through joint
attention, temporal consistency and motion alignment are learned implicitly, without
optical flow estimation or handcrafted temporal regularization.

\paragraph{Positional Anchoring for Motion Preservation.}
Preserving fine-grained spatial structure and long-range motion correspondence is a
central challenge in video style transfer.
To this end, we introduce a positional anchoring mechanism that enforces a hard
alignment prior between the source and target videos.

Specifically, we assign identical spatiotemporal positional encodings to source video
tokens $T_c$ and target (denoising) video tokens $T_d$ that correspond to the same
temporal index and spatial location.
By construction, tokens at the same $(t, h, w)$ position in the source and target
videos share the same positional identity throughout the diffusion process.
This anchors each denoising token to its corresponding content location, effectively
suppressing spatial drift and preserving motion consistency without relying on
optical flow estimation or heuristic correspondence propagation.

This design can be interpreted as enforcing a shared spatiotemporal identity between
source and target tokens, i.e.,
\begin{equation}
(t, h, w)_{T_c} \equiv (t, h, w)_{T_d},
\end{equation}
indicating that both videos are defined on the same spatiotemporal lattice.

\begin{figure*}[t]
\centering
\includegraphics[width=\linewidth]{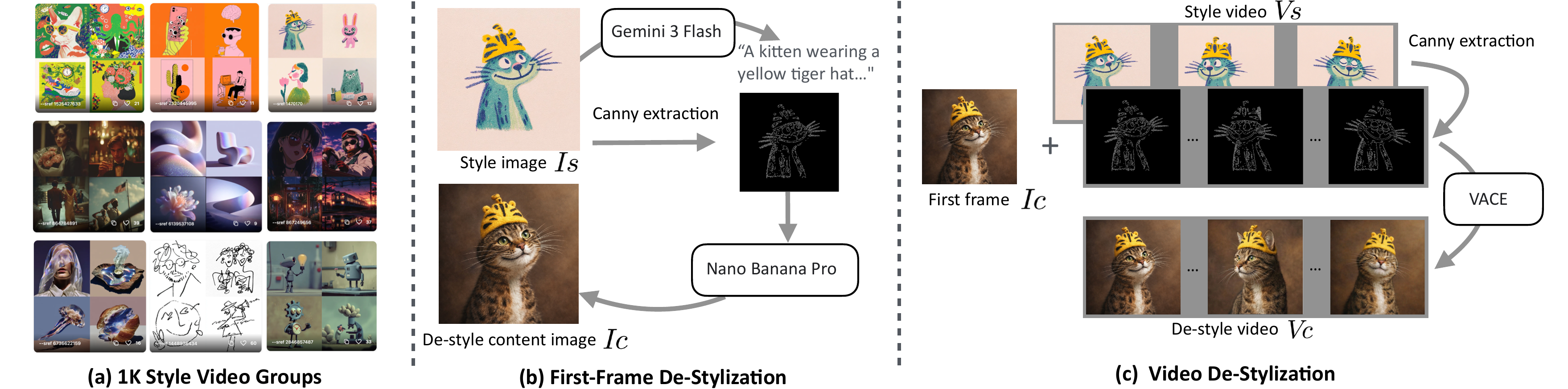}
\captionof{figure}{Overview of the VISTA-1000 data construction pipeline. An inverse synthesis process is used to derive clean, motion-consistent videos from stylized ones, enabling the construction of triplet data that explicitly separates style, content, and motion.}
\vspace*{-1em}
\label{fig:data_constructure}
\end{figure*}

\subsection{Style Representation and Injection}
\label{sec:style_adapter}

While triplet supervision provides strong alignment signals between style and motion,
the scale and diversity of video style transfer data remain inherently limited compared
to large-scale image corpora.
In particular, learning robust and fine-grained style representations solely from video
data may hinder generalization across diverse visual appearances.
To address this limitation, we introduce a lightweight \emph{Style Adapter} that
explicitly leverages image-level pretraining to provide strong appearance priors.

\paragraph{Two-Stage Training of the Style Adapter.}
The Style Adapter is trained in two stages.
In the first stage, we pretrain the adapter on large-scale natural images to learn
generalizable style representations.
Specifically, we adopt \textbf{SigLIP} \cite{tschannen2025siglip}as the image encoder and extract fine-grained
visual features from the last transformer layer before global pooling, which preserves
local texture, color composition, and detailed appearance patterns.
Following the IP-Adapter paradigm, we treat the pretrained Diffusion Transformer
backbone as an image generation model and optimize the adapter using a self-conditioned
denoising objective, where the conditioning image and the denoising target correspond
to the same image. This stage enables the adapter to capture transferable style priors without relying on
task-specific video supervision. In the second stage, the pretrained Style Adapter is integrated into the shared-context diffusion framework and jointly fine-tuned on video style transfer triplets.
Through this adaptation, the adapter becomes style-aware in the video domain while
retaining the generalization benefits of image-level pretraining.

\paragraph{Shared-query Dual-KV Fusion.}
Given a style reference image $I_s$, we extract fine-grained style features using a pretrained SigLIP encoder. We select the feature map from the last transformer block \emph{before} global pooling:
\begin{equation}
F_s = \mathrm{SigLIP}(I_s) \in \mathbb{R}^{N \times D_s},
\end{equation}
where $N$ is the number of patches and $D_s$ is the feature dimension.
At layer $\ell$, a lightweight Style Adapter maps $F_s$ to style-specific keys and values via learnable projectors:
\begin{equation}
K^{(s)}_\ell = F_s W^{(s)}_{K,\ell},\qquad
V^{(s)}_\ell = F_s W^{(s)}_{V,\ell}.
\end{equation}
Let $Q_\ell$ be the query derived from the diffusion latent tokens, and $\{K^{(mm)}_\ell, V^{(mm)}_\ell\}$ be the keys and values from the backbone multi-modal context (e.g., text and video conditions).
We compute the backbone attention and a parallel style-only attention using the same query $Q_\ell$, then fuse their outputs \emph{before} the shared output projection:
\begin{equation}
\begin{aligned}
\tilde{O}_\ell &=
\mathrm{Attn}(Q_\ell, K^{(mm)}_\ell, V^{(mm)}_\ell)
+ \alpha_\ell\, \mathrm{Attn}(Q_\ell, K^{(s)}_\ell, V^{(s)}_\ell),
\end{aligned}
\label{eq:style_fusion}
\end{equation}
where $\alpha_\ell$ is a learnable scalar controlling the injection strength. This design allows the model to query appearance cues from $I_s$ while preserving the motion structure learned in the backbone.

\subsection{VISTA-1000 Dataset Construction}
\label{sec:dataset}

We construct VISTA-1000 as a tuple dataset containing a style reference image $I_s$, a de-stylized content image $I_c$, a de-stylized content video $V_c$, and a stylized target video $V_s$, as shown in Figure~\ref{fig:data_constructure}. 
The target videos $V_s$ are collected from Sref, a curated platform of Midjourney-defined artistic styles~\cite{midjourney_home}. 
Professional designers manually select $1{,}000$ high-quality styles with strong visual consistency, and the corresponding stylized videos are used as targets. Rather than using forward stylization, we build training data through inverse de-stylization. 
For each $V_s$, Nano-Banana~\cite{google_nanobanana_doc} first generates a de-stylized first-frame image $I_c$ that preserves content structure while removing style-specific appearance. 
We then condition VACE on $I_c$ and Canny edge maps extracted from $V_s$ to synthesize a de-stylized content video $V_c$ aligned with $V_s$ in motion and layout. 
After human filtering, each pair $(V_c,V_s)$ is combined with a style reference image $I_s$ from the same style group but with different semantics. 
With four videos per style, this yields about $12{,}000$ training tuples and promotes disentanglement among style, content, and motion.

\subsection{Efficient Causal Distillation}
After training the bidirectional VISTA teacher, we accelerate inference through a video-to-video causal distillation pipeline adapted from the ideas of diffusion forcing~\cite{chen2024diffusion} and Self-Forcing~\cite{huang2025self}. 
In the first stage, the student is initialized from the teacher and converted from full spatiotemporal attention to block-causal attention with cross-chunk KV caching, enabling autoregressive video generation. 
Inspired by diffusion forcing, we expose the student to heterogeneous per-frame noise levels, so that clean past chunks and noisy current chunks can coexist during training, better matching the streaming inference setting. 
This stage still follows a teacher-forcing input policy, since the noisy inputs are constructed from ground-truth stylized videos rather than the student's own predictions. 
In the second stage, inspired by the on-policy rollout strategy of Self-Forcing, we further fine-tune the student under its own denoising trajectory. 
Different from text-to-video self-forcing, our video-to-video style transfer setting provides a paired stylized target for each source video and style reference, allowing direct ground-truth regression without matching a frozen teacher or using adversarial/score-distillation losses. 
After distillation, the sampling process is compressed to four denoising steps, substantially improving inference speed while trading off a certain degree of visual quality.

\begin{figure*}[t]
\centering
\animategraphics[width=1.0\linewidth]{5}{sec/image/fig1_v3_q85/fig1-}{00001}{00010}
\captionof{figure}{Generation results of VISTA. Readers can click and play the video clips in this figure using {\color{red}\textbf{Adobe Acrobat}}.}
\vspace*{-1em}
\label{fig:result}
\end{figure*}

\section{Experiment}
\label{gen_inst}

\subsection{Experiment Setting}

\noindent \textbf{Training Details.}  
Our experiments are conducted on the pretrained WAN 2.2 Diffusion Transformer backbone under the video-to-video stylization setting.  
We train our model at a resolution of 768$\times$768 with 81-frame sequences.  
Following recent diffusion fine-tuning practices, we employ the LoRA-based adaptation strategy with a LoRA rank of 256, batch size 16, and learning rate $2\times10^{-4}$.  
The model is fine-tuned for 20,000 steps using the AdamW optimizer with a cosine learning rate schedule.  
During training, we adopt mixed precision and gradient checkpointing for efficiency.  
All experiments are conducted on a cluster of 8 NVIDIA A100 GPUs.The generation results are shown in  Fig. \ref{fig:result}.



\noindent \textbf{Baseline Methods.}
We compare VISTA with representative video stylization and generation systems, including Stylemaster~\cite{ye2025stylemaster}and Anyv2v~\cite{ku2024anyv2v}(open-source), as well as Kling o1~\cite{team2025kling} and Runway Alpha4~\cite{runway_gen4_page}(closed-source commercial APIs).
We exclude models such as Google VEO3, as they do not support video-to-video inputs required for video stylization.
In addition, we include hybrid baselines that combine image stylization with video generation or video-to-video models.
Specifically, we adopt a strong state-of-the-art image stylization method (Instant Style) to stylize the first frame, and then use this stylized frame together with a video-to-video control model (VACE) to generate the full video.
For the VACE branch, we use Canny edge maps extracted from the source video as structural guidance.

\noindent \textbf{Benchmarks.}
We evaluate all methods on two benchmarks: the VISTA Benchmark and the Public Benchmark.
The VISTA Benchmark consists of 50 held-out videos from our dataset, disjoint from the training set, while the Public Benchmark contains 50 videos collected from established video editing benchmarks.
Each video includes 81 frames.
All methods are evaluated under identical settings, using one reference image and the same preprocessing and normalization pipelines to ensure fair comparison.

To provide a more comprehensive and convincing evaluation, we additionally introduce a Public Benchmark for supplementary experiments. This benchmark consists of 50 video–style pairs, where the input videos are drawn from the DAVIS benchmark \citep{bertrand2023test} and the reference style images are randomly selected from WikiArt \citep{shen2018neural}.

\noindent \textbf{Evaluation Metrics.}
Video style transfer is a multi-condition problem that requires preserving the content structure and motion dynamics of the source video while faithfully adopting the appearance of the reference style image and maintaining high visual quality.
We therefore evaluate performance from four aspects:
\textbf{style consistency}, measured by CSD\citep{kim2023collaborative} and CLIP Image Score between the generated video and the reference style image;
\textbf{content consistency}, measured by LPIPS and CLIP Image Score between the generated video and the input source video;
\textbf{motion consistency}, measured by CoTracker3 \citep{karaev2025cotracker3} through motion trajectory comparison between input and output videos;
and \textbf{aesthetic quality}, evaluated using VBench Aesthetic and Imaging scores \citep{huang2025vbench++}.

\begin{figure*}[htb]
\centering
\animategraphics[width=\linewidth]{5}{figures/fig2/fig2_frame_}{01}{10}
\caption{Comparison results show that our method consistently outperforms baseline approaches in style consistency, content preservation, motion coherence, and overall visual quality, and remains competitive with state-of-the-art commercial models. Readers can click and play the video clips in this figure using {\color{red}\textbf{Adobe Acrobat}}.}
\label{fig:qualitative}
\end{figure*}

\begin{table*}[htb]
\centering
\caption{Quantitative comparison on two benchmarks. $\uparrow$ indicates higher is better, $\downarrow$ lower is better.
CLIP$_S$ and CLIP$_C$ denote CLIP-based style similarity and content similarity, respectively. We evaluate motion consistency between the source and stylized videos using Motion Fidelity computed from CoTracker3 trajectories.}
\label{tab:quantitative}
\resizebox{\textwidth}{!}{
\begin{tabular}{lccccccc ccccccc}
\toprule
& \multicolumn{7}{c}{VISTA Bench} 
& \multicolumn{7}{c}{Public Bench} \\
\cmidrule(lr){2-8}\cmidrule(lr){9-15}

\textbf{Method}
& \multicolumn{2}{c}{\textbf{Style}} 
& \multicolumn{2}{c}{\textbf{Content}} 
& \multicolumn{1}{c}{\textbf{Motion}} 
& \multicolumn{2}{c}{\textbf{Aesthetic}}
& \multicolumn{2}{c}{\textbf{Style}} 
& \multicolumn{2}{c}{\textbf{Content}} 
& \multicolumn{1}{c}{\textbf{Motion}} 
& \multicolumn{2}{c}{\textbf{Aesthetic}} \\

\cmidrule(lr){2-3}\cmidrule(lr){4-5}\cmidrule(lr){6-6}\cmidrule(lr){7-8}
\cmidrule(lr){9-10}\cmidrule(lr){11-12}\cmidrule(lr){13-13}\cmidrule(lr){14-15}

& CSD $\uparrow$ & CLIP$_S$ $\uparrow$
& LPIPS $\downarrow$ & CLIP$_C$ $\uparrow$
& Fidelity $\uparrow$
& Aesthetic $\uparrow$ & Imaging $\uparrow$
& CSD $\uparrow$ & CLIP$_S$ $\uparrow$
& LPIPS $\downarrow$ & CLIP$_C$ $\uparrow$
& Fidelity $\uparrow$
& Aesthetic $\uparrow$ & Imaging $\uparrow$ \\
\midrule

Stylemaster
& 0.377 & 0.632 & 0.624 & 0.740 & 0.171 & 0.606 & 0.701
& 0.393 & 0.583 & 0.697 & 0.710 & 0.161 & 0.455 & 0.670 \\

Anyv2v
& 0.389 & 0.640 & 0.616 & 0.688 & 0.229 & 0.547 & 0.681
& 0.343 & 0.578 & 0.718 & 0.577 & 0.196 & 0.419 & 0.623 \\

VACE
& 0.460 & 0.671 & 0.568 & 0.769 & 0.397 & 0.593 & 0.706
& 0.403 & 0.599 & 0.653 & 0.651 & 0.325 & 0.452 & 0.648 \\

Kling o1
& 0.448 & \textbf{0.671} & 0.559 & 0.800 & 0.351 & 0.653 & 0.730
& 0.416 & 0.584 & \textbf{0.617} & \textbf{0.756} & 0.265 & 0.512 & \textbf{0.694} \\

Runway
& 0.460 & 0.663 & 0.572 & 0.788 & \textbf{0.415} & 0.635 & 0.714
& 0.414 & 0.586 & 0.638 & 0.726 & \textbf{0.356} & 0.489 & 0.655 \\

Ours
& \textbf{0.467} & 0.664 & \textbf{0.532} & \textbf{0.802} & 0.380 & \textbf{0.682} & \textbf{0.746}
& \textbf{0.421} & \textbf{0.600} & 0.634 & 0.723 & 0.317 & \textbf{0.544} & 0.689 \\

Ours (distill)
& 0.453 & 0.658 & 0.530 & 0.795 & 0.352 & 0.664 & 0.708
& 0.413 & 0.588 & 0.614 & 0.719 & 0.279 & 0.527 & 0.651 \\

\bottomrule
\end{tabular}
}
\end{table*}

\subsection{Experiment Results}
\label{sec:qualitative}

Figure~\ref{fig:result} and Figure~\ref{fig:qualitative} presents qualitative comparisons between VISTA and representative open-source and commercial baselines.
Overall, VISTA produces more visually pleasing stylized videos, with stronger style adherence, cleaner textures, and better preserved content structures.
Compared with existing video-to-video stylization or editing methods, our method better maintains foreground identity, scene layout, and temporally consistent stylized details across frames.The transferred styles are consistently reflected in color tone, texture patterns, and overall artistic appearance, while the source content remains recognizable and structurally stable.

In contrast, baseline methods often suffer from weak or incomplete style transfer, noisy textures, over-smoothed appearances, drifting stylized patterns, or distorted object boundaries under complex motion and deformation.
Commercial systems such as Runway and Kling can generate strong results in some cases, but may still show inconsistent style strength or partial content changes.
These comparisons demonstrate that VISTA more reliably applies the reference style while preserving motion-aligned content, highlighting the benefit of learning video stylization from paired triplet supervision.

\subsection{Quantitative Evaluation and Efficiency Analysis}
\label{sec:quantitative}

Table~\ref{tab:quantitative} reports quantitative comparisons on the VISTA and Public benchmarks.
On the \textit{VISTA Benchmark}, VISTA achieves the best overall performance, with the highest style consistency (CSD), the lowest LPIPS, strong motion fidelity, and the best aesthetic and imaging scores.
These results indicate that our method achieves a favorable balance among style adherence, content preservation, temporal stability, and perceptual quality.
On the \textit{Public Benchmark}, VISTA remains highly competitive and achieves best or near-best results across key metrics, particularly in style consistency and visual quality, while maintaining motion fidelity comparable to strong commercial APIs such as Runway and Kling.
Overall, the quantitative results validate the advantage of triplet-based supervision and data-driven in-context learning over heuristic or handcrafted consistency mechanisms.

We further evaluate the inference efficiency of the distilled student model.
On a single NVIDIA H200 GPU, the bidirectional teacher takes about 580 seconds to generate an 81-frame video, whereas the distilled four-step student only requires about 6 seconds.
Although this acceleration introduces some degradation in visual quality, the student model provides a practical fast-preview mode for video stylization and suggests its potential for future real-time video generation and editing applications.

\begin{table}[t]
\centering
\scriptsize
\setlength{\tabcolsep}{3pt}
\renewcommand{\arraystretch}{0.92}

\begin{minipage}[t]{0.49\linewidth}
\centering
\caption{Ablation study results. Removing either the Style Adapter or the style tokens degrades stylistic performance.}
\label{tab:ablation}
\resizebox{\linewidth}{!}{
\begin{tabular}{lccc ccc}
\toprule
& \multicolumn{3}{c}{VISTA Bench} & \multicolumn{3}{c}{Public Bench} \\
\cmidrule(lr){2-4}\cmidrule(lr){5-7}
\textbf{Method} 
& CSD $\uparrow$ & Aes. $\uparrow$ & Imag. $\uparrow$
& CSD $\uparrow$ & Aes. $\uparrow$ & Imag. $\uparrow$ \\
\midrule
w/o Style Tokens   
& 0.406 & 0.633 & 0.702 
& 0.382 & 0.498 & 0.652 \\
w/o Style Adapter  
& 0.377 & 0.590 & 0.654 
& 0.395 & 0.445 & 0.621 \\
\textbf{Full Model} 
& \textbf{0.467} & \textbf{0.682} & \textbf{0.746} 
& \textbf{0.421} & \textbf{0.544} & \textbf{0.689} \\
\bottomrule
\end{tabular}
}
\end{minipage}
\hfill
\begin{minipage}[t]{0.49\linewidth}
\centering
\caption{Ablation on positional anchoring.}
\label{tab:ablation_posanch}
\resizebox{\linewidth}{!}{
\begin{tabular}{lccc ccc}
\toprule
& \multicolumn{3}{c}{VISTA Bench} & \multicolumn{3}{c}{Public Bench} \\
\cmidrule(lr){2-4}\cmidrule(lr){5-7}
\textbf{Method}
& \multicolumn{2}{c}{\textbf{Content}}
& \multicolumn{1}{c}{\textbf{Motion}}
& \multicolumn{2}{c}{\textbf{Content}}
& \multicolumn{1}{c}{\textbf{Motion}} \\
\cmidrule(lr){2-3}\cmidrule(lr){4-4}
\cmidrule(lr){5-6}\cmidrule(lr){7-7}
& LPIPS $\downarrow$ & CLIP$_{\text{c}}$ $\uparrow$
& Fidel. $\uparrow$
& LPIPS $\downarrow$ & CLIP$_{\text{c}}$ $\uparrow$
& Fidel. $\uparrow$ \\
\midrule
w/o Pos. Anchoring
& 0.546 & 0.780 & 0.371
& 0.651 & 0.707 & 0.311 \\
\textbf{Full Model}
& \textbf{0.532} & \textbf{0.802} & \textbf{0.380}
& \textbf{0.634} & \textbf{0.723} & \textbf{0.317} \\
\bottomrule
\end{tabular}
}
\end{minipage}

\vspace{-0.5em}
\end{table}

\subsection{Ablation Study}
\label{sec:ablation}

We conduct ablation studies on the VISTA benchmark to analyze the contribution of key components in VISTA, including
(1) \textbf{w/o Style-In-Context Tokens},
(2) \textbf{w/o Style Adapter}, and
(3) the full \textbf{VISTA} model.

\begin{figure}[t]
\centering
\captionsetup{font=small}

\begin{minipage}[t]{0.4\linewidth}
  \centering
  \animategraphics[width=\linewidth]{5}{sec/image/fig3_v4_q85/fig3-}{00001}{00010}
  \captionof{figure}{Ablation study results. Video clips can be played in {\color{red}\textbf{Adobe Acrobat}}.}
  \label{fig:ablation}
\end{minipage}
\hfill
\begin{minipage}[t]{0.58\linewidth}
  \centering
  \includegraphics[width=\linewidth]{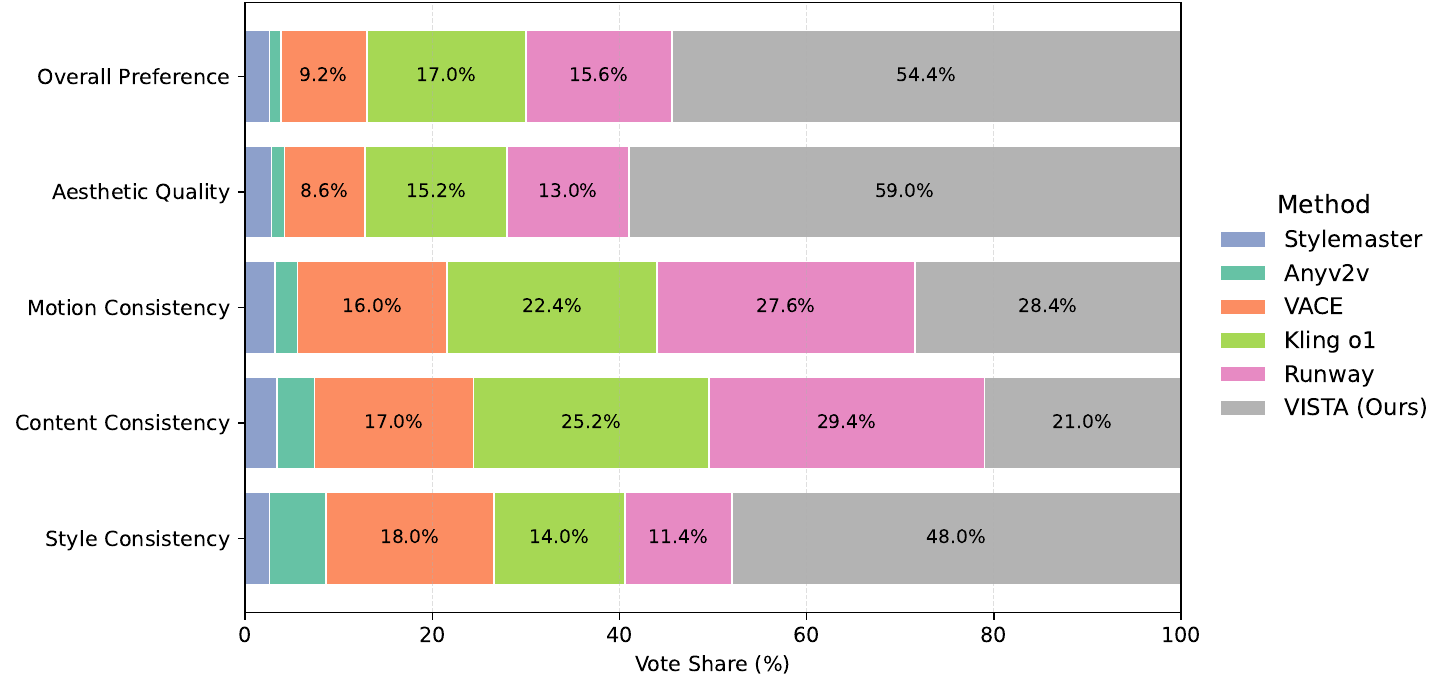}
  \captionof{figure}{User study results. VISTA achieves the highest preference in overall quality, aesthetics, and style consistency.}
  \label{fig:user_study}
\end{minipage}

\end{figure}

As shown in Table~\ref{tab:ablation}, removing Style-In-Context Tokens weakens the interaction between style and video context, leading to degraded stylization quality.
Removing the Style Adapter limits the model’s ability to capture fine-grained appearance cues from a single reference image.
The full model achieves the best performance, demonstrating that shared-context conditioning and the Style Adapter play complementary roles. The generated results are shown in the Figure \ref{fig:ablation}. We further evaluate the proposed positional anchoring mechanism in Table~\ref{tab:ablation_posanch}.
Removing positional anchoring results in consistent drops in both motion and content consistency on the VISTA and Public benchmarks, confirming its importance for stabilizing motion alignment and preserving content structure.

\subsection{User Study}
We conduct a user study to compare VISTA with baseline methods from a human preference perspective (Fig.~\ref{fig:user_study}).
All methods are anonymized and evaluated through an online questionnaire.
Participants are shown the source video and stylized results, and asked to select their preferred output based on five criteria: overall preference, aesthetics, motion consistency, content consistency, and style consistency.
We randomly sample 20 comparison groups and collect 25 valid anonymous responses, reporting results as vote percentages. VISTA is consistently preferred in overall quality, aesthetics, and style consistency, while remaining competitive with strong commercial baselines such as Runway in motion and content consistency.
These results align with our quantitative evaluation and confirm that VISTA better matches human judgments on visual quality and temporal stability.

\section{Conclusion}
\label{sec:conclusion}
In this work, we presented VISTA, a data-driven framework for temporally consistent video style transfer. Our key insight is that reliable video stylization should be learned from structured triplet supervision rather than enforced by handcrafted temporal heuristics. To enable this, we constructed a high-quality paired triplet dataset with 1,000 styles and 12,000 triplets, where clean and stylized videos share aligned motion and content, while style references provide appearance cues with different semantics. Built on this dataset, we trained a diffusion transformer with shared-context conditioning and introduced a lightweight Style Adapter with two-stage training to better leverage large-scale natural image priors. Extensive evaluations demonstrate improved style fidelity, temporal stability, and content preservation across diverse styles and motions, highlighting the effectiveness of combining structured data with dedicated model design.

\bibliographystyle{plainnat}
\bibliography{references}

\begin{thebibliography}{60}
\providecommand{\natexlab}[1]{#1}
\providecommand{\url}[1]{\texttt{#1}}
\expandafter\ifx\csname urlstyle\endcsname\relax
  \providecommand{\doi}[1]{doi: #1}\else
  \providecommand{\doi}{doi: \begingroup \urlstyle{rm}\Url}\fi

\bibitem[Batifol et~al.(2025)Batifol, Blattmann, Boesel, Consul, Diagne, Dockhorn, English, English, Esser, Kulal, et~al.]{batifol2025flux}
Stephen Batifol, Andreas Blattmann, Frederic Boesel, Saksham Consul, Cyril Diagne, Tim Dockhorn, Jack English, Zion English, Patrick Esser, Sumith Kulal, et~al.
\newblock Flux. 1 kontext: Flow matching for in-context image generation and editing in latent space.
\newblock \emph{arXiv e-prints}, pages arXiv--2506, 2025.

\bibitem[Bertrand et~al.(2023)Bertrand, Kordopatis~Zilos, Kalantidis, and Tolias]{bertrand2023test}
Juliette Bertrand, Giorgos Kordopatis~Zilos, Yannis Kalantidis, and Giorgos Tolias.
\newblock Test-time training for matching-based video object segmentation.
\newblock \emph{Advances in Neural Information Processing Systems}, 36:\penalty0 20918--20941, 2023.

\bibitem[Chen et~al.(2024)Chen, Mart{\'\i}~Mons{\'o}, Du, Simchowitz, Tedrake, and Sitzmann]{chen2024diffusion}
Boyuan Chen, Diego Mart{\'\i}~Mons{\'o}, Yilun Du, Max Simchowitz, Russ Tedrake, and Vincent Sitzmann.
\newblock Diffusion forcing: Next-token prediction meets full-sequence diffusion.
\newblock \emph{Advances in Neural Information Processing Systems}, 37:\penalty0 24081--24125, 2024.

\bibitem[Chen et~al.(2026)Chen, Mao, Song, Gu, and Ma]{chen2026edittransfer++}
Lan Chen, Qi~Mao, Yiren Song, Yuchao Gu, and Siwei Ma.
\newblock Edittransfer++: Toward faithful and efficient visual-prompt-guided image editing.
\newblock \emph{arXiv preprint arXiv:2605.07455}, 2026.

\bibitem[Feng et~al.(2024)Feng, Weng, Wang, Yuan, Bao, Luo, Chen, and Guo]{feng2024ccedit}
Ruoyu Feng, Wenming Weng, Yanhui Wang, Yuhui Yuan, Jianmin Bao, Chong Luo, Zhibo Chen, and Baining Guo.
\newblock Ccedit: Creative and controllable video editing via diffusion models.
\newblock In \emph{Proceedings of the IEEE/CVF Conference on Computer Vision and Pattern Recognition}, pages 6712--6722, 2024.

\bibitem[Gao et~al.(2025)Gao, Zhou, Du, Zhang, and Gan]{gao2025adaworld}
Shenyuan Gao, Siyuan Zhou, Yilun Du, Jun Zhang, and Chuang Gan.
\newblock Adaworld: Learning adaptable world models with latent actions.
\newblock \emph{arXiv preprint arXiv:2503.18938}, 2025.

\bibitem[Gao et~al.(2020)Gao, Li, Yin, and Yang]{gao2020fast}
Wei Gao, Yijun Li, Yihang Yin, and Ming-Hsuan Yang.
\newblock Fast video multi-style transfer.
\newblock In \emph{Proceedings of the IEEE/CVF winter conference on applications of computer vision}, pages 3222--3230, 2020.

\bibitem[Geyer et~al.(2023)Geyer, Bar-Tal, Bagon, and Dekel]{geyer2023tokenflow}
Michal Geyer, Omer Bar-Tal, Shai Bagon, and Tali Dekel.
\newblock Tokenflow: Consistent diffusion features for consistent video editing.
\newblock \emph{arXiv preprint arXiv:2307.10373}, 2023.

\bibitem[Gong et~al.(2025)Gong, Song, Li, Li, and Zhang]{gong2025relationadapter}
Yan Gong, Yiren Song, Yicheng Li, Chenglin Li, and Yin Zhang.
\newblock Relationadapter: Learning and transferring visual relation with diffusion transformers.
\newblock \emph{arXiv preprint arXiv:2506.02528}, 2025.

\bibitem[{Google AI for Developers}(2026)]{google_nanobanana_doc}
{Google AI for Developers}.
\newblock Nano banana image generation, 2026.
\newblock URL \url{https://ai.google.dev/gemini-api/docs/image-generation}.
\newblock Last updated 2026-01-22 UTC.

\bibitem[Guo et~al.(2025)Guo, Zeng, Song, Zhang, Zhang, and Liu]{guo2025any2anytryon}
Hailong Guo, Bohan Zeng, Yiren Song, Wentao Zhang, Chuang Zhang, and Jiaming Liu.
\newblock Any2anytryon: Leveraging adaptive position embeddings for versatile virtual clothing tasks.
\newblock \emph{arXiv preprint arXiv:2501.15891}, 2025.

\bibitem[Guo et~al.(2023)Guo, Yang, Rao, Liang, Wang, Qiao, Agrawala, Lin, and Dai]{guo2023animatediff}
Yuwei Guo, Ceyuan Yang, Anyi Rao, Zhengyang Liang, Yaohui Wang, Yu~Qiao, Maneesh Agrawala, Dahua Lin, and Bo~Dai.
\newblock Animatediff: Animate your personalized text-to-image diffusion models without specific tuning.
\newblock \emph{arXiv preprint arXiv:2307.04725}, 2023.

\bibitem[Huang et~al.(2017)Huang, Wang, Luo, Ma, Jiang, Zhu, Li, and Liu]{huang2017real}
Haozhi Huang, Hao Wang, Wenhan Luo, Lin Ma, Wenhao Jiang, Xiaolong Zhu, Zhifeng Li, and Wei Liu.
\newblock Real-time neural style transfer for videos.
\newblock In \emph{Proceedings of the IEEE conference on computer vision and pattern recognition}, pages 783--791, 2017.

\bibitem[Huang et~al.(2023)Huang, Chen, Liu, Shen, Zhao, and Zhou]{huang2023composer}
Lianghua Huang, Di~Chen, Yu~Liu, Yujun Shen, Deli Zhao, and Jingren Zhou.
\newblock Composer: Creative and controllable image synthesis with composable conditions.
\newblock \emph{arXiv preprint arXiv:2302.09778}, 2023.

\bibitem[Huang and Belongie(2017)]{huang2017arbitrary}
Xun Huang and Serge Belongie.
\newblock Arbitrary style transfer in real-time with adaptive instance normalization.
\newblock In \emph{Proceedings of the IEEE international conference on computer vision}, pages 1501--1510, 2017.

\bibitem[Huang et~al.(2025{\natexlab{a}})Huang, Li, He, Zhou, and Shechtman]{huang2025self}
Xun Huang, Zhengqi Li, Guande He, Mingyuan Zhou, and Eli Shechtman.
\newblock Self forcing: Bridging the train-test gap in autoregressive video diffusion.
\newblock \emph{arXiv preprint arXiv:2506.08009}, 2025{\natexlab{a}}.

\bibitem[Huang et~al.(2025{\natexlab{b}})Huang, Zhang, Xu, He, Yu, Dong, Ma, Chanpaisit, Si, Jiang, Wang, Chen, Chen, Wang, Lin, Qiao, and Liu]{huang2025vbench++}
Ziqi Huang, Fan Zhang, Xiaojie Xu, Yinan He, Jiashuo Yu, Ziyue Dong, Qianli Ma, Nattapol Chanpaisit, Chenyang Si, Yuming Jiang, Yaohui Wang, Xinyuan Chen, Ying-Cong Chen, Limin Wang, Dahua Lin, Yu~Qiao, and Ziwei Liu.
\newblock {VBench++}: Comprehensive and versatile benchmark suite for video generative models.
\newblock \emph{IEEE Transactions on Pattern Analysis and Machine Intelligence}, 2025{\natexlab{b}}.
\newblock \doi{10.1109/TPAMI.2025.3633890}.

\bibitem[Jiang et~al.(2024)Jiang, Mao, Pan, Han, and Zhang]{jiang2024scedit}
Zeyinzi Jiang, Chaojie Mao, Yulin Pan, Zhen Han, and Jingfeng Zhang.
\newblock Scedit: Efficient and controllable image diffusion generation via skip connection editing.
\newblock In \emph{Proceedings of the IEEE/CVF conference on computer vision and pattern Recognition}, pages 8995--9004, 2024.

\bibitem[Jiang et~al.(2025)Jiang, Han, Mao, Zhang, Pan, and Liu]{jiang2025vace}
Zeyinzi Jiang, Zhen Han, Chaojie Mao, Jingfeng Zhang, Yulin Pan, and Yu~Liu.
\newblock Vace: All-in-one video creation and editing.
\newblock \emph{arXiv preprint arXiv:2503.07598}, 2025.

\bibitem[Karaev et~al.(2025)Karaev, Makarov, Wang, Neverova, Vedaldi, and Rupprecht]{karaev2025cotracker3}
Nikita Karaev, Yuri Makarov, Jianyuan Wang, Natalia Neverova, Andrea Vedaldi, and Christian Rupprecht.
\newblock Cotracker3: Simpler and better point tracking by pseudo-labelling real videos.
\newblock In \emph{Proceedings of the IEEE/CVF International Conference on Computer Vision}, pages 6013--6022, 2025.

\bibitem[Kim et~al.(2023)Kim, Lee, Choi, Jeong, Sohn, and Shin]{kim2023collaborative}
Subin Kim, Kyungmin Lee, June~Suk Choi, Jongheon Jeong, Kihyuk Sohn, and Jinwoo Shin.
\newblock Collaborative score distillation for consistent visual synthesis.
\newblock \emph{arXiv preprint arXiv:2307.04787}, 2023.

\bibitem[Ku et~al.(2024)Ku, Wei, Ren, Yang, and Chen]{ku2024anyv2v}
Max Ku, Cong Wei, Weiming Ren, Harry Yang, and Wenhu Chen.
\newblock Anyv2v: A tuning-free framework for any video-to-video editing tasks.
\newblock \emph{arXiv preprint arXiv:2403.14468}, 2024.

\bibitem[Li et~al.(2017)Li, Fang, Yang, Wang, Lu, and Yang]{li2017universal}
Yijun Li, Chen Fang, Jimei Yang, Zhaowen Wang, Xin Lu, and Ming-Hsuan Yang.
\newblock Universal style transfer via feature transforms.
\newblock \emph{Advances in neural information processing systems}, 30, 2017.

\bibitem[Lin et~al.(2025)Lin, Jiang, Yang, Zheng, Liang, Zhang, and Liu]{lin2025omnihuman}
Gaojie Lin, Jianwen Jiang, Jiaqi Yang, Zerong Zheng, Chao Liang, Yuan Zhang, and Jingtuo Liu.
\newblock Omnihuman-1: Rethinking the scaling-up of one-stage conditioned human animation models.
\newblock In \emph{Proceedings of the IEEE/CVF International Conference on Computer Vision}, pages 13847--13858, 2025.

\bibitem[Liu et~al.(2025)Liu, Ouyang, Lou, and Song]{liu2025omnirefiner}
Yaoli Liu, Ziheng Ouyang, Shengtao Lou, and Yiren Song.
\newblock Omnirefiner: Reinforcement-guided local diffusion refinement.
\newblock \emph{arXiv preprint arXiv:2511.19990}, 2025.

\bibitem[Ma et~al.(2024)Ma, Liu, Wang, Pan, He, Yuan, Zeng, Cai, Shum, Liu, et~al.]{ma2024followyouremoji}
Yue Ma, Hongyu Liu, Hongfa Wang, Heng Pan, Yingqing He, Junkun Yuan, Ailing Zeng, Chengfei Cai, Heung-Yeung Shum, Wei Liu, et~al.
\newblock Follow-your-emoji: Fine-controllable and expressive freestyle portrait animation.
\newblock In \emph{SIGGRAPH Asia 2024 Conference Papers}, pages 1--12, 2024.

\bibitem[Ma et~al.(2025{\natexlab{a}})Ma, He, Wang, Wang, Shen, Qi, Ying, Cai, Li, Shum, et~al.]{ma2025follow}
Yue Ma, Yingqing He, Hongfa Wang, Andong Wang, Leqi Shen, Chenyang Qi, Jixuan Ying, Chengfei Cai, Zhifeng Li, Heung-Yeung Shum, et~al.
\newblock Follow-your-click: Open-domain regional image animation via motion prompts.
\newblock In \emph{Proceedings of the AAAI Conference on Artificial Intelligence}, volume~39, pages 6018--6026, 2025{\natexlab{a}}.

\bibitem[Ma et~al.(2025{\natexlab{b}})Ma, He, Wang, Wang, Shen, Qi, Ying, Cai, Li, Shum, et~al.]{ma2025followyourclick}
Yue Ma, Yingqing He, Hongfa Wang, Andong Wang, Leqi Shen, Chenyang Qi, Jixuan Ying, Chengfei Cai, Zhifeng Li, Heung-Yeung Shum, et~al.
\newblock Follow-your-click: Open-domain regional image animation via motion prompts.
\newblock In \emph{Proceedings of the AAAI Conference on Artificial Intelligence}, volume~39, pages 6018--6026, 2025{\natexlab{b}}.

\bibitem[{Midjourney}(2026)]{midjourney_home}
{Midjourney}.
\newblock Midjourney, 2026.
\newblock URL \url{https://www.midjourney.com/}.

\bibitem[Ouyang et~al.(2025)Ouyang, Song, Liu, Zhu, Hou, Cheng, and Shou]{ouyang2025consistency}
Ziheng Ouyang, Yiren Song, Yaoli Liu, Shihao Zhu, Qibin Hou, Ming-Ming Cheng, and Mike~Zheng Shou.
\newblock The consistency critic: Correcting inconsistencies in generated images via reference-guided attentive alignment.
\newblock \emph{arXiv preprint arXiv:2511.20614}, 2025.

\bibitem[Pan et~al.(2017)Pan, Qiu, Yao, Li, and Mei]{pan2017create}
Yingwei Pan, Zhaofan Qiu, Ting Yao, Houqiang Li, and Tao Mei.
\newblock To create what you tell: Generating videos from captions.
\newblock In \emph{Proceedings of the 25th ACM international conference on Multimedia}, pages 1789--1798, 2017.

\bibitem[Peebles and Xie(2023)]{peebles2023scalable}
William Peebles and Saining Xie.
\newblock Scalable diffusion models with transformers.
\newblock In \emph{Proceedings of the IEEE/CVF international conference on computer vision}, pages 4195--4205, 2023.

\bibitem[Qi et~al.(2023)Qi, Cun, Zhang, Lei, Wang, Shan, and Chen]{qi2023fatezero}
Chenyang Qi, Xiaodong Cun, Yong Zhang, Chenyang Lei, Xintao Wang, Ying Shan, and Qifeng Chen.
\newblock Fatezero: Fusing attentions for zero-shot text-based video editing.
\newblock In \emph{Proceedings of the IEEE/CVF International Conference on Computer Vision}, pages 15932--15942, 2023.

\bibitem[Qin et~al.(2024)Qin, Li, Tang, Chua, and Zhuang]{qin2024instructvid2vid}
Bosheng Qin, Juncheng Li, Siliang Tang, Tat-Seng Chua, and Yueting Zhuang.
\newblock Instructvid2vid: Controllable video editing with natural language instructions.
\newblock In \emph{2024 IEEE International Conference on Multimedia and Expo (ICME)}, pages 1--6. IEEE, 2024.

\bibitem[Ruder et~al.(2016)Ruder, Dosovitskiy, and Brox]{ruder2016artistic}
Manuel Ruder, Alexey Dosovitskiy, and Thomas Brox.
\newblock Artistic style transfer for videos.
\newblock In \emph{German conference on pattern recognition}, pages 26--36. Springer, 2016.

\bibitem[{Runway}(2025)]{runway_gen4_page}
{Runway}.
\newblock Introducing runway gen-4, 2025.
\newblock URL \url{https://runwayml.com/research/introducing-runway-gen-4}.

\bibitem[Schaldenbrand et~al.(2022)Schaldenbrand, Liu, and Oh]{schaldenbrand2022styleclipdraw}
Peter Schaldenbrand, Zhixuan Liu, and Jean Oh.
\newblock Styleclipdraw: Coupling content and style in text-to-drawing translation.
\newblock \emph{arXiv preprint arXiv:2202.12362}, 2022.

\bibitem[Shen et~al.(2018)Shen, Yan, and Zeng]{shen2018neural}
Falong Shen, Shuicheng Yan, and Gang Zeng.
\newblock Neural style transfer via meta networks.
\newblock In \emph{Proceedings of the IEEE Conference on Computer Vision and Pattern Recognition}, pages 8061--8069, 2018.

\bibitem[Song et~al.(2025{\natexlab{a}})Song, Song, Peng, Gao, and Shou]{song2025worldwander}
Quanjian Song, Yiren Song, Kelly Peng, Yuan Gao, and Mike~Zheng Shou.
\newblock Worldwander: Bridging egocentric and exocentric worlds in video generation.
\newblock \emph{arXiv preprint arXiv:2511.22098}, 2025{\natexlab{a}}.

\bibitem[Song(2022)]{song2022cliptexture}
Yiren Song.
\newblock Cliptexture: Text-driven texture synthesis.
\newblock In \emph{Proceedings of the 30th ACM International Conference on Multimedia}, pages 5468--5476, 2022.

\bibitem[Song and Zhang(2022)]{song2022clipfont}
Yiren Song and Yuxuan Zhang.
\newblock Clipfont: Text guided vector wordart generation.
\newblock In \emph{BMVC}, page 543, 2022.

\bibitem[Song et~al.(2023)Song, Shao, Chen, Zhang, Jing, and Li]{song2023clipvg}
Yiren Song, Xuning Shao, Kang Chen, Weidong Zhang, Zhongliang Jing, and Minzhe Li.
\newblock Clipvg: Text-guided image manipulation using differentiable vector graphics.
\newblock In \emph{Proceedings of the AAAI conference on artificial intelligence}, volume~37, pages 2312--2320, 2023.

\bibitem[Song et~al.(2024)Song, Huang, Yao, Ye, Ci, Liu, Zhang, and Shou]{processpainter}
Yiren Song, Shijie Huang, Chen Yao, Xiaojun Ye, Hai Ci, Jiaming Liu, Yuxuan Zhang, and Mike~Zheng Shou.
\newblock Processpainter: Learn painting process from sequence data.
\newblock \emph{arXiv preprint arXiv:2406.06062}, 2024.

\bibitem[Song et~al.(2025{\natexlab{b}})Song, Liu, and Shou]{song2025makeanything}
Yiren Song, Cheng Liu, and Mike~Zheng Shou.
\newblock Makeanything: Harnessing diffusion transformers for multi-domain procedural sequence generation.
\newblock \emph{arXiv preprint arXiv:2502.01572}, 2025{\natexlab{b}}.

\bibitem[Song et~al.(2025{\natexlab{c}})Song, Liu, and Shou]{song2025omniconsistency}
Yiren Song, Cheng Liu, and Mike~Zheng Shou.
\newblock Omniconsistency: Learning style-agnostic consistency from paired stylization data.
\newblock \emph{arXiv preprint arXiv:2505.18445}, 2025{\natexlab{c}}.

\bibitem[Team et~al.(2025)Team, Chen, Ci, Du, Feng, Gai, Guo, Han, He, He, et~al.]{team2025kling}
Kling Team, Jialu Chen, Yuanzheng Ci, Xiangyu Du, Zipeng Feng, Kun Gai, Sainan Guo, Feng Han, Jingbin He, Kang He, et~al.
\newblock Kling-omni technical report.
\newblock \emph{arXiv preprint arXiv:2512.16776}, 2025.

\bibitem[Tschannen et~al.(2025)Tschannen, Gritsenko, Wang, Naeem, Alabdulmohsin, Parthasarathy, Evans, Beyer, Xia, Mustafa, et~al.]{tschannen2025siglip}
Michael Tschannen, Alexey Gritsenko, Xiao Wang, Muhammad~Ferjad Naeem, Ibrahim Alabdulmohsin, Nikhil Parthasarathy, Talfan Evans, Lucas Beyer, Ye~Xia, Basil Mustafa, et~al.
\newblock Siglip 2: Multilingual vision-language encoders with improved semantic understanding, localization, and dense features.
\newblock \emph{arXiv preprint arXiv:2502.14786}, 2025.

\bibitem[Wan et~al.(2025)Wan, Wang, Ai, Wen, Mao, Xie, Chen, Yu, Zhao, Yang, et~al.]{wan2025wan}
Team Wan, Ang Wang, Baole Ai, Bin Wen, Chaojie Mao, Chen-Wei Xie, Di~Chen, Feiwu Yu, Haiming Zhao, Jianxiao Yang, et~al.
\newblock Wan: Open and advanced large-scale video generative models.
\newblock \emph{arXiv preprint arXiv:2503.20314}, 2025.

\bibitem[Wang et~al.(2024{\natexlab{a}})Wang, Spinelli, Wang, Bai, Qin, and Chen]{wang2024instantstyle}
Haofan Wang, Matteo Spinelli, Qixun Wang, Xu~Bai, Zekui Qin, and Anthony Chen.
\newblock Instantstyle: Free lunch towards style-preserving in text-to-image generation.
\newblock \emph{arXiv preprint arXiv:2404.02733}, 2024{\natexlab{a}}.

\bibitem[Wang et~al.(2024{\natexlab{b}})Wang, Bai, Wang, Qin, Chen, Li, Tang, and Hu]{wang2024instantid}
Qixun Wang, Xu~Bai, Haofan Wang, Zekui Qin, Anthony Chen, Huaxia Li, Xu~Tang, and Yao Hu.
\newblock Instantid: Zero-shot identity-preserving generation in seconds.
\newblock \emph{arXiv preprint arXiv:2401.07519}, 2024{\natexlab{b}}.

\bibitem[Wang et~al.(2020)Wang, Xu, Zhang, Wang, and Liu]{wang2020consistent}
Wenjing Wang, Jizheng Xu, Li~Zhang, Yue Wang, and Jiaying Liu.
\newblock Consistent video style transfer via compound regularization.
\newblock In \emph{Proceedings of the AAAI conference on artificial intelligence}, volume~34, pages 12233--12240, 2020.

\bibitem[Wang et~al.(2023{\natexlab{a}})Wang, Yuan, Zhang, Chen, Wang, Zhang, Shen, Zhao, and Zhou]{wang2023videocomposer}
Xiang Wang, Hangjie Yuan, Shiwei Zhang, Dayou Chen, Jiuniu Wang, Yingya Zhang, Yujun Shen, Deli Zhao, and Jingren Zhou.
\newblock Videocomposer: Compositional video synthesis with motion controllability.
\newblock \emph{Advances in Neural Information Processing Systems}, 36:\penalty0 7594--7611, 2023{\natexlab{a}}.

\bibitem[Wang et~al.(2023{\natexlab{b}})Wang, Wang, Xie, Qi, Shan, Wang, and Luo]{wang2023styleadapter}
Zhouxia Wang, Xintao Wang, Liangbin Xie, Zhongang Qi, Ying Shan, Wenping Wang, and Ping Luo.
\newblock Styleadapter: A unified stylized image generation model.
\newblock \emph{arXiv preprint arXiv:2309.01770}, 2023{\natexlab{b}}.

\bibitem[Xu et~al.(2024)Xu, Zhang, Liew, Yan, Liu, Zhang, Feng, and Shou]{xu2024magicanimate}
Zhongcong Xu, Jianfeng Zhang, Jun~Hao Liew, Hanshu Yan, Jia-Wei Liu, Chenxu Zhang, Jiashi Feng, and Mike~Zheng Shou.
\newblock Magicanimate: Temporally consistent human image animation using diffusion model.
\newblock In \emph{Proceedings of the IEEE/CVF Conference on Computer Vision and Pattern Recognition}, pages 1481--1490, 2024.

\bibitem[Yan et~al.(2025)Yan, Ma, Zou, Chen, Chen, and Zhang]{yan2025eedit}
Zexuan Yan, Yue Ma, Chang Zou, Wenteng Chen, Qifeng Chen, and Linfeng Zhang.
\newblock Eedit: Rethinking the spatial and temporal redundancy for efficient image editing.
\newblock \emph{arXiv preprint arXiv:2503.10270}, 2025.

\bibitem[Ye et~al.(2023)Ye, Zhang, Liu, Han, and Yang]{ye2023ip}
Hu~Ye, Jun Zhang, Sibo Liu, Xiao Han, and Wei Yang.
\newblock Ip-adapter: Text compatible image prompt adapter for text-to-image diffusion models.
\newblock \emph{arXiv preprint arXiv:2308.06721}, 2023.

\bibitem[Ye et~al.(2025)Ye, Huang, Wang, Wan, Zhang, and Luo]{ye2025stylemaster}
Zixuan Ye, Huijuan Huang, Xintao Wang, Pengfei Wan, Di~Zhang, and Wenhan Luo.
\newblock Stylemaster: Stylize your video with artistic generation and translation.
\newblock In \emph{Proceedings of the Computer Vision and Pattern Recognition Conference}, pages 2630--2640, 2025.

\bibitem[Zhang et~al.(2023)Zhang, Rao, and Agrawala]{zhang2023adding}
Lvmin Zhang, Anyi Rao, and Maneesh Agrawala.
\newblock Adding conditional control to text-to-image diffusion models.
\newblock In \emph{Proceedings of the IEEE/CVF international conference on computer vision}, pages 3836--3847, 2023.

\bibitem[Zhang et~al.(2024)Zhang, Wei, ZHANG, Zuo, Tian, et~al.]{zhang2305controlvideo}
Yabo Zhang, Yuxiang Wei, XIAOPENG ZHANG, Wangmeng Zuo, Qi~Tian, et~al.
\newblock Controlvideo: Training-free controllable text-to-video generation.
\newblock In \emph{International Conference on Learning Representations}, volume 2024, pages 54441--54461, 2024.

\bibitem[Zheng et~al.(2024)Zheng, Peng, Yang, Shen, Li, Liu, Zhou, Li, and You]{zheng2024open}
Zangwei Zheng, Xiangyu Peng, Tianji Yang, Chenhui Shen, Shenggui Li, Hongxin Liu, Yukun Zhou, Tianyi Li, and Yang You.
\newblock Open-sora: Democratizing efficient video production for all.
\newblock \emph{arXiv preprint arXiv:2412.20404}, 2024.

\end{thebibliography}







\end{document}